\pdfoutput=1

\documentclass[11pt]{article}

\usepackage{EMNLP2023}

\usepackage{times}
\usepackage{latexsym}

\usepackage{amsmath}
\usepackage{amsfonts}
\usepackage{mathtools}
\usepackage{amsthm}
\usepackage{mathrsfs}
\usepackage{amssymb}
\usepackage{booktabs}
\usepackage{adjustbox}

\usepackage[T1]{fontenc}

\usepackage[utf8]{inputenc}

\usepackage{microtype}

\usepackage{inconsolata}

\usepackage{amsmath, amsfonts, amsthm, fullpage, amssymb, color, graphicx, enumitem, mathtools, tikz, thmtools, float}

\usetikzlibrary{graphs,graphs.standard}

\usepackage{svg}

\def \[#1\] {
  \begin{align*}
  #1
  \end{align*}
}
\newcommand{\eq}[1]{\begin{align*}#1\end{align*}}

\newtheorem{lemma}{Lemma}
\theoremstyle{definition}

\makeatletter  
\def\@endtheorem{\qed\endtrivlist\@endpefalse } 
\makeatother
\newtheoremstyle{proof}
  {0pt}
  {\topsep}
  {}
  {0pt}
  {\itshape}
  {.}
  { }
  {\thmname{#1}\thmnumber{ #2}\textnormal{\thmnote{ (#3)}}}
\theoremstyle{proof}
\newtheorem*{prf}{Proof}

\newcommand{\ce}{\coloneqq}
\newcommand{\set}[1]{\{#1\}}

\newcommand{\brk}[1]{\left[ #1 \right]}

\newcommand{\simm}{\mathrm{\ sim \ }}

\title{Referral Augmentation for Zero-Shot Information Retrieval}

\author{Michael Tang, Shunyu Yao, John Yang, Karthik Narasimhan \\
  Department of Computer Science, Princeton University\\ 
  \texttt{\{mwtang, shunyuy, jy1682, karthikn\}@princeton.edu} \\}

\begin{document}
\maketitle
\begin{abstract}
We propose Referral-Augmented Retrieval (RAR), a simple technique that concatenates document indices with 
\textit{referrals}, i.e. text from other documents that cite or link to the given document, to provide significant performance gains for zero-shot information retrieval. The key insight behind our method is that referrals provide a more complete, multi-view representation of a document, much like incoming page links in algorithms like PageRank provide a comprehensive idea of a webpage's importance.
RAR works with both sparse and dense retrievers, and outperforms generative text expansion techniques such as DocT5Query~\cite{doc2query} and Query2Doc~\cite{query2doc} — a 37\% and 21\% absolute improvement on ACL paper retrieval Recall@10 — while also eliminating expensive model training and inference. 
We also analyze different methods for multi-referral aggregation and show that RAR enables up-to-date information retrieval without re-training.\footnote{Code: \url{https://github.com/michaelwilliamtang/referral-augment}}

\end{abstract}

\section{Introduction}

Zero-shot information retrieval, a task in which both test queries and corpora are inaccessible at training time, closely mimics real-world deployment settings where the distribution of text changes over time and the system needs to continually adapt to new queries and documents. Prior work~\cite{beir} finds that without access to training on in-domain query-document pairs or task-specific document relations, most dense models 
dramatically underperform simple sparse models like BM25, pointing to poor generalization. At the same time, sparse models struggle to reconcile different surface forms, leading to the so-called \textit{lexical gap} between queries and documents in different tasks.

\begin{figure}[t]
\begin{centering}
\includegraphics[width=0.4\textwidth]{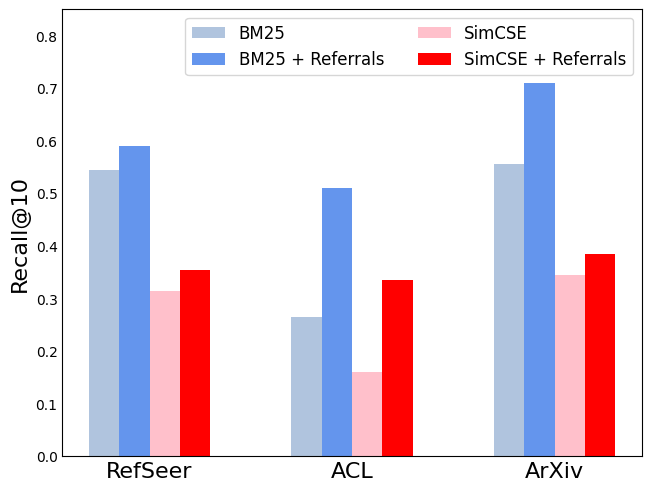}
\caption{ Our referral augmentation method improves zero-shot document retrieval across a variety of models and datasets.
}
\label{fig_teaser}
\end{centering}
\end{figure}

\begin{figure*}[t]
\centering
\includegraphics[width=\textwidth]{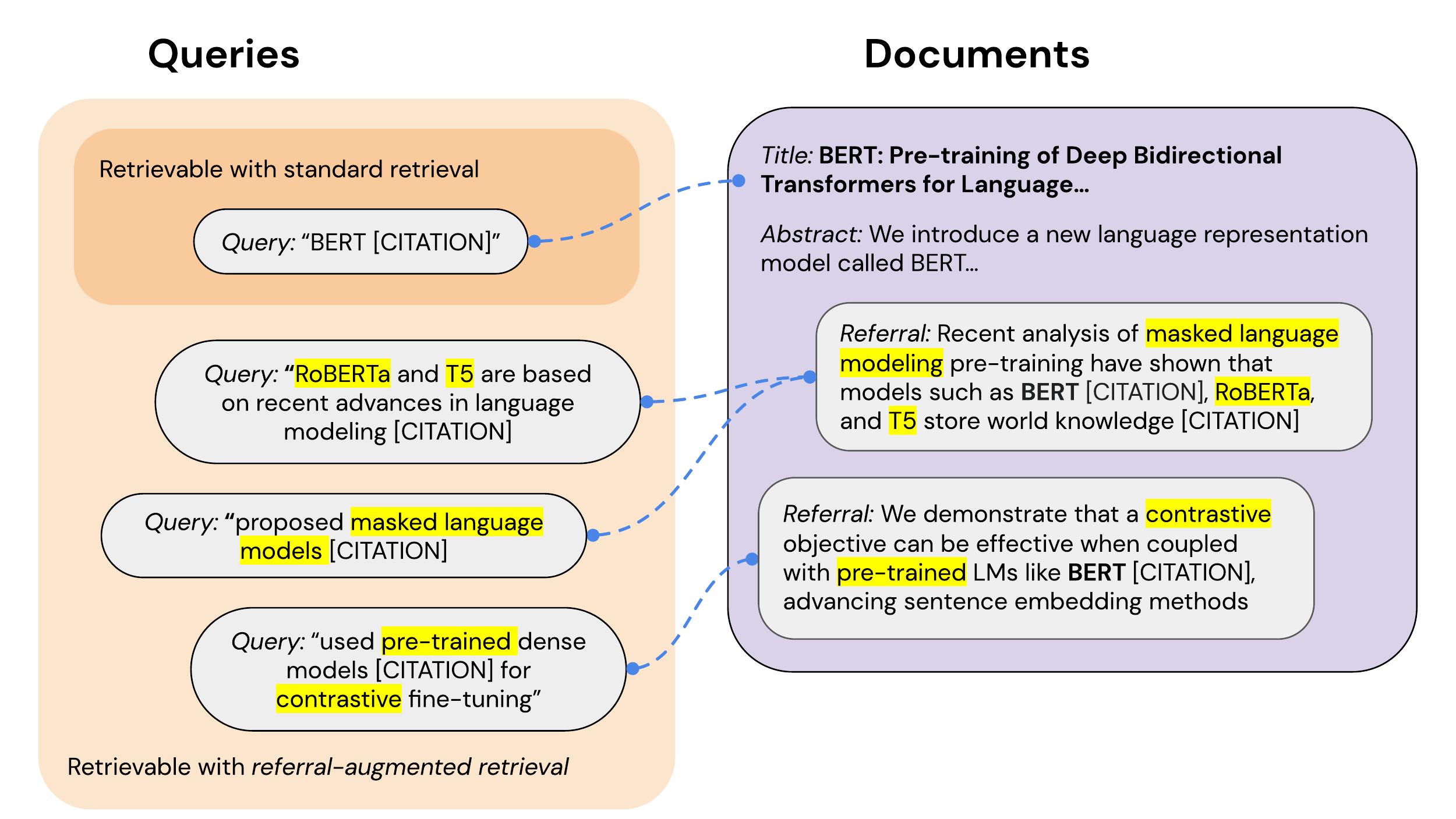}
\caption{\label{fig_overview} Illustration of the Referral-Augmented Retrieval (RAR) process. RAR augments text from documents that refer to the original document into its index (right), which allows it to correctly retrieve the target document for a wider range of queries (left) compared to standard methods. This example uses text around citations as queries, from the citation recommendation task~\cite{lcr_nianlong}.}
\end{figure*}

\begin{figure*}
\begin{centering}
\includegraphics[width=\textwidth]{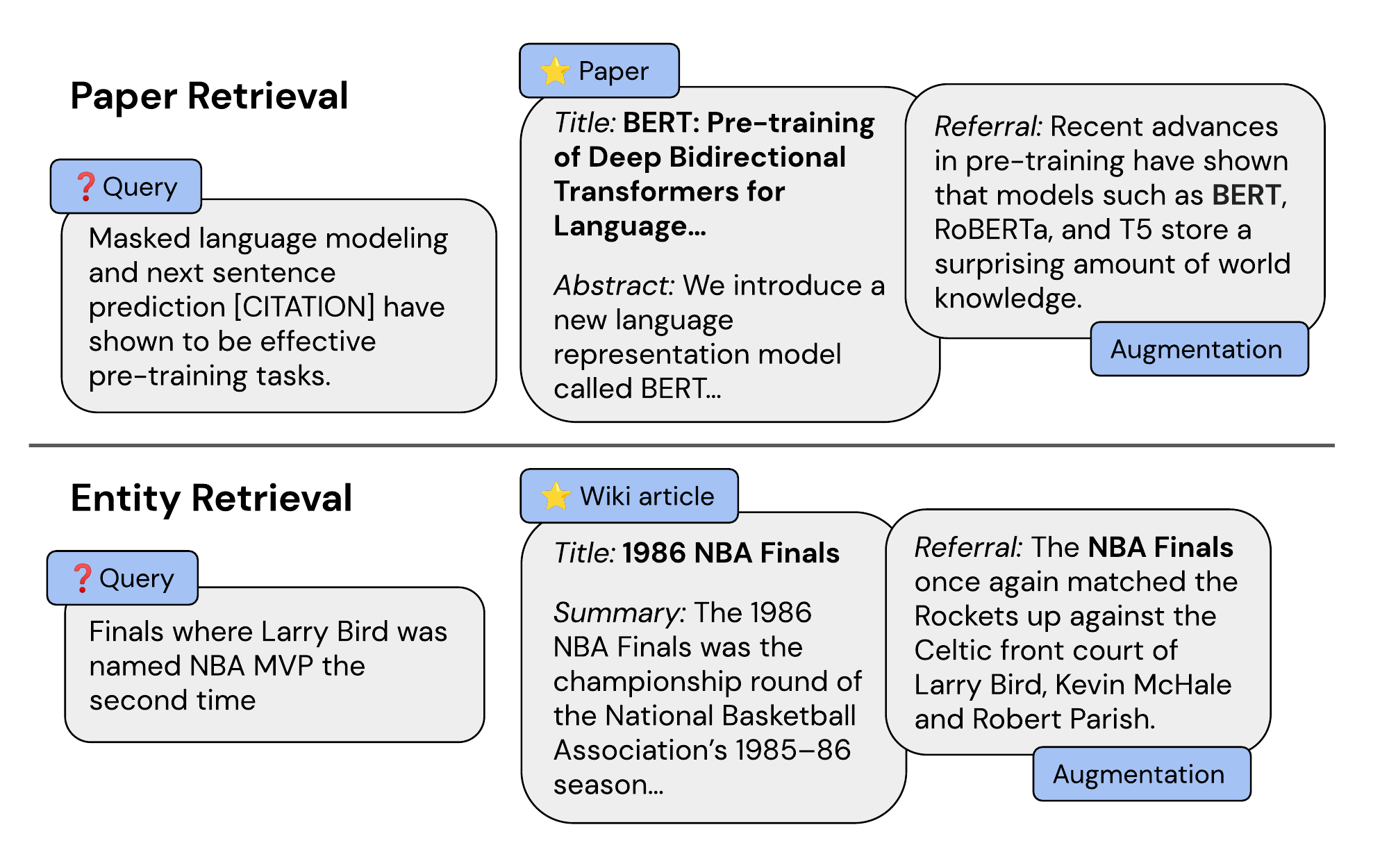}
\caption{We evaluate referral augmentation on zero-shot paper retrieval, retrieving papers given masked in-text citations, (top) and entity retrieval, retrieving wiki articles on each titular entity given free text queries about the entity (bottom).}
\label{fig_paper_entity_retrieval}
\end{centering}
\end{figure*}

While the zero-shot setting lacks query-document pairs, our key insight is to leverage intra-document relations that provide multiple views of the same information to provide a more comprehensive representations of the concepts in a document. We propose Referral-Augmented Retrieval (RAR), a simple technique that augments the text of each document in a retrieval index with passages from other documents that contain citations or hyperlinks to it. This use of intra-document information is reminiscent of Google's BackRub and PageRank algorithms --- while they leverage the number of intra-doc linkages to estimate a document's importance \cite{thesearch}, we leverage the content of linkages as examples of how they are usually \textit{referred}, and thus might be similarly \textit{retrieved}. 

We evaluate our method on both a paper retrieval setting on a corpus of academic papers and an entity retrieval setting on a corpus of Wikipedia pages, and find that RAR significantly improves zero-shot retrieval performance for both sparse and dense models. For instance, RAR outperforms generative text expansion techniques such as DocT5Query~\cite{doc2query} and Query2Doc~\cite{query2doc} by up to 37\% and 21\% Recall@10, respectively, on ACL paper retrieval from the S2ORC corpus~\cite{lo-etal-2022-semantic}. Moreover, RAR's augmentation occurs entirely at indexing time and hence allows for a training-free method to update a retrieval system with new views of existing documents (e.g., a trending news story that causes users to search for a public figure by the name of the scandal they were in). Our method also scales well as the number of referrals increases and is easy to update.

We also observe interesting connections between RAR and prior query or document 
expansion techniques~\cite{doc2query, hyde, query2doc}.
Text expansion techniques effectively surface \textit{hard positives}, passages that are very lexically different but semantically equivalent, including conceptual transformations (e.g., mapping a claim to a piece of contradictory evidence), the addition of new information, and alternative formulations with different word choice or scope. While some of these transformations are theoretically learnable, existing dense retrievers are often not robust to them, so explicitly augmenting documents and queries with their equivalent counterparts significantly improves the encoded representations. As an added bonus, the text-to-text nature of these hard positive pairs allows them to be both model-agnostic and interpretable. This observation motivates further research into improving retrieval not by training a more expressing encoder, but \textit{by simply discovering more hard positives}. 

\section{Method}
\subsection{Preliminaries}

Formally, given a set of queries $Q$ and documents $D$, retrieval can be described as the task of learning a similarity function $\simm(q,d)$ between a query $q \in Q$ and a document $d \in D$, where top-$k$ retrieval is equivalent to finding the ordered tuple $(d_1,...,d_k)$ where
\eq{
\simm(q,d_1) &\ge ... \ge \simm(q,d_k) \\ &\ge \simm(q,d) \quad \forall d \notin \set{d_1,...,d_k}
}
For dense models, similarity is typically computed as the dot product between the encodings of queries, where the encoder is shared:
\eq{
\simm(q,d) \ce f(q) \cdot f(d)
}
We can formally define a hard positive as a pair of highly relevant passages $\set{x_1, x_2}$ that should be mapped to the same point in embedding space, which in effect imposes a correction on top of a given encoder $f$ where $f(x_1) \ne f(x_2)$.
\subsection{Connections to related work} 
Under our framework, the query generation technique DocT5Query \cite{doc2query} corresponds to generating $\ell$ hard positive pairs $(\set{q_i(d), d})_{i=1}^\ell$ for each $d \in D$, each of which is a question about that document generated by a T5 model. For inference, they apply BM25 on the expanded documents $\tilde d \ce [d, q_1(d),...,q_\ell(d)]$ where $[\cdot,\cdot]$ denotes concatenation.

Similarly, the hypothetical document generation techniques HyDE and Query2Doc \cite{hyde, query2doc} correspond to generating $\ell$ hard positive pairs $(\set{q, d_i(q)}_{i=1}^\ell$ \textit{at inference time} for a given query $q$, each of which is a hypothetical document generated by InstructGPT to answer the query. For inference, HyDE uses the mean dense encoding between each hypothetical document $\tilde f(q) \ce \frac{1}{\ell+1} \brk{q+\sum_i d_i(q)}$, whereas Query2Doc applies BM25 on the augmented query $\tilde q \ce [q, d_1(q),...,d_\ell(q)]$ (they use $\ell=1$, and repeat the original query $q$ a total of $n=5$ times to emphasize its relative importance).

\subsection{Referrals}
We directly use document-to-document relations in the corpus metadata as hard positives, obtaining up to $\ell$ pairs $(\set{q_i(d), d})_{i=1}^\ell$ for each $d \in D$ which are sentences in other documents containing citations or hyperlinks to the current document $d$. We experiment with three different referral integration methods: \label{aggregation}
\begin{enumerate}
    \item \textbf{Concatenation}: $\tilde d \ce [d, q_1(d),...,q_\ell(d)]$
    \item \textbf{Mean} $\tilde f(d) \ce \frac{1}{\ell+1} \brk{f(d)+\sum_i f(q_i(d))}$
    \item \textbf{Shortest path} $\tilde {\simm} (q, d) \ce \min \set{\simm(q,d),(\simm(q,q_i(d)))_{i=1}^\ell}$
\end{enumerate}
We find in Section \ref{integration} that for sparse models, concatenation performs the best, while for dense models, mean aggregation performs the best, although shortest path achieves the best top 1 accuracy (Recall@1) since it preserves the high granularity of separate referrals, and use these settings when reporting overall results.

\section{Experiments}
\subsection{Setup}
\paragraph{Paper retrieval} Paper retrieval is the task of retrieving papers most likely to be cited in a given passage. We partition a corpus of papers into disjoint candidate and evaluation sets — papers in the candidate set represent older, known papers we want to retrieve, while papers in the evaluation set represent newer papers whose body text may cite those older papers, each citation inducing a retrieval task with a ground truth. Following the classic setup of \textit{local citation recommendation} (LCR) \cite{lcr_nianlong}, we represent each candidate paper via its concatenated title and abstract, and construct a query from each sentence in an evaluation papers referencing a candidate paper (with the citation masked). To evaluate the effects of augmenting a candidate document at indexing time, we compile referrals consisting of citing sentences in \textit{other candidate papers}.

We compare performance with and without augmentation on ACL and ArXiv papers from the S2ORC corpus \cite{s2orc}, as well as the open-domain RefSeer corpus. ACL and ArXiv paper retrieval tasks were partitioned such that papers published in 2018 or before comprised the candidate set, and papers in 2019 comprised the evaluation set, filtering to only include candidate papers that were cited at least once. In-text citations were masked out in both queries and referrals; queries consisted of just the citing sentence, whereas referrals used a 200-token window centered around the masked in-text citation. Documents were augmented with a uniform random sample of up to $\ell=30$ referrals.

\paragraph{Entity retrieval} Entity retrieval is the task of retrieving the most relevant entities from a knowledge base given a text query. We evaluate on the DBPedia entity retrieval task, which represents each entity (associated with a Wikipedia page) via its concatenated name and summary, and contains freeform text queries. To augment a candidate document, we compile referrals consisting of sentences from the pages of other entities that link to the the document. We used the 2017 English Wikipedia dump preprocessed with WikiExtractor \cite{Wikiextractor} and extract hyperlinks via a HTML parser, again including a random sample of up to 30 referrals per document.

\paragraph{Models} For the retriever, we use BM25 \cite{bm25} as a sparse baseline and SimCSE \cite{simcse} and DPR \cite{dpr}, contrastively fine-tuned BERT encoders, as dense baselines. We also evaluate on BM25 + CE, which adds a cross-encoder to the BM25 model \cite{minilm} and was found to be the best-performing zero-shot retriever from the BEIR evaluation \cite{beir}. For paper retrieval, we also evaluate the effect of using referrals with Specter \cite{specter}, a domain-specific encoder pre-trained and fine-tuned on scientific text.

\begin{table*}
\centering
\begin{tabular}{lccc} \toprule
   & \multicolumn{1}{p{2.5cm}}{\centering RefSeer}
   & \multicolumn{1}{p{2.5cm}}{\centering ACL}
   & \multicolumn{1}{p{2.5cm}}{\centering ArXiv} \\ \toprule
   && \textit{Recall@10 \quad (Recall@1)}\\
   \midrule
   BM25 & 0.545 (0.260) & 0.265 (0.115) & 0.555 (0.335) \\
   \quad + referrals & \textbf{0.590} (\textbf{0.335}) & \textbf{0.505} (\textbf{0.200}) & \textbf{0.710}  (\textbf{0.430}) \\ \midrule
   SimCSE & 0.315 (0.095) & 0.160 (0.065) & 0.345 (\textbf{0.140}) \\
   \quad + referrals & \textbf{0.355} (\textbf{0.155}) & \textbf{0.355} (\textbf{0.115}) & \textbf{0.385} (0.120) \\
   \bottomrule

\end{tabular}

\caption{\label{table-lcr} Paper retrieval results with citation referrals. RAR greatly improves paper retrieval performance for both sparse and dense models on all metrics, sometimes doubling the absolute performance.}

\end{table*}

\begin{table*}
\centering
\begin{tabular}{lccc} \toprule
   & \multicolumn{1}{p{1.8cm}}{\centering \textit{nDCG@1}}
   & \multicolumn{1}{p{1.8cm}}{\centering \textit{nDCG@10}}
   & \multicolumn{1}{p{1.8cm}}{\centering \textit{Recall@10}} \\
   \midrule
   BM25 & 0.4030 & 0.2739 & \textbf{0.1455} \\
   \quad + referrals & \textbf{0.4851} & \textbf{0.2799} & 0.1348 \\
   \midrule
   BM25 + CE & 0.4254 & 0.3282 & 0.1798 \\
   \quad + referrals & \textbf{0.4478} & \textbf{0.3283} & \textbf{0.1949} \\
   \midrule
   DPR & 0.3350 & 0.2559 & 0.1562 \\
   \quad + referrals & \textbf{0.3538} & \textbf{0.2610} & \textbf{0.1612} \\
   \bottomrule
\end{tabular}
\caption{\label{table-beir-2} Entity retrieval results with hyperlink referrals, on the DBPedia task. RAR improves entity retrieval performance on both sparse and dense models.}
\end{table*}

\subsection{Results}
\paragraph{Paper retrieval}

From Table \ref{table-lcr}, we see that a retriever augmented with referrals outperforms the base retriever for all sparse and dense models, with significant improvement on both Recall@1 and Recall@10 on all datasets (including an extremely large 100\% improvement on ACL) for BM25 + referrals compared to regular BM25. We see that alongside surfacing more relevant information to increase recall, referrals also greatly increase the specificity to generate much better top-1 retrieved candidates, pointing to the fact that referring citations referencing a paper are often more clear, concise, and well-specified than the abstract of the paper itself. 

\paragraph{Entity retrieval}
We evaluate model performance with and without referrals in Table \ref{table-beir-2}. We see that referrals again significantly elevate performance for both sparse and dense models across the board. The gain is particularly large for nDCG@1, which we hypothesize is due to the occasionally extremely high similarity of referring sentences with some queries.

We note that hyperlink referrals do not increase performance as much as the respective citation referrals on the paper retrieval task, suggesting that linking sentences may be less consistent and less directly informative than citing ones. Intuitively, different citations of a given scientific work are typically similar in spirit, while the relevance relations implied by different hyperlinks may be more tangential. However, this is not necessarily a fair comparison, as the Wikipedia-based query and corpus distributions also vary much more and encompass more diverse fields of knowledge.

\section{Analysis}
\subsection{Referrals outperform other augmentations}
\begin{table*}
\centering
\begin{tabular}{lccc} \toprule
   & \multicolumn{1}{p{2cm}}{\centering \textit{Recall@1}}
   & \multicolumn{1}{p{2cm}}{\centering \textit{MRR@10}}
   & \multicolumn{1}{p{2cm}}{\centering \textit{Recall@10}} \\ \toprule
   BM25 & 0.13 & 0.177 & 0.29 \\ 
   \quad + referrals & \textbf{0.35} & \textbf{0.4088} & \textbf{0.53} \\ \midrule
   \quad + DocT5Query & 0.0 & 0.036 & 0.155 \\
   \quad + DocT5Query + referrals & 0.345 & 0.4022 & 0.525 \\ \midrule
   \quad + Query2Doc & 0.14 & 0.1940 & 0.32 \\
   \quad + Query2Doc + referrals & \textbf{0.38} & \textbf{0.4279} & 0.52 \\
   \bottomrule
\end{tabular}
\caption{\label{table-stack} Paper retrieval, referrals vs. other augmentation techniques (Recall@10). We bold the best result on any single augmentation strategy, as well as any results on stacked augmentations that show further gains over that single augmentation. Overall, we find that referrals greatly outperform other augmentation techniques, and further that referrals can stack with Query2Doc to achieve even better performance.}
\end{table*}

In Table \ref{table-stack}, we show that referral augmentation strongly outperforms query and document augmentation techniques exemplified by DocT5Query and Query2Doc. Generative models like DocT5Query fail to capture the more complex text distribution on domains like scientific papers and generate qualitatively nonsensical or trivial queries, whereas referrals leverage gold quality reformulations of the paper directly from document-to-document links.

\subsection{Referral aggregation methods} \label{integration}

Aggregating dense representations is a well-known problem \cite{fusion, rawgnn, aggretriever}, and is usually resolved via concatenation or taking a sum or average. We propose three such methods: text concatenation, mean representation, and shortest path (details in section \ref{aggregation}), which we will denote by referrals \textsubscript{concat}, referrals \textsubscript{mean}, referrals \textsubscript{sp}. Note that BM25 does not support mean aggregation since it does not yield vector embeddings.

In particular, we add the shortest path method as a novel option in order to take advantage of different referrals representing distinct views of a given document that should not necessarily be aggregated as a single mean embedding —

while citations are fairly consistent, hyperlinks to a given article sometimes focus on unrelated aspects of its content (e.g. referencing a famous painting by its painter vs. by its host museum) which may be best represented by different locations in query space.

\paragraph{Results} We evaluate them in Table \ref{table-aggregation} and find that text concatenation performs the best for BM25 but poorly for SimCSE, which we hypothesize is due to the fact that repetition and concatenation of text improves the approximation of a target query (inverse term frequency) distribution for BM25, but results in 
a distorted dense representation since dense models approach text sequentially and in particular a long string of referring sentences in a row is very much out of their training distribution.

For dense models, mean and shortest path aggregation performs the best for Recall@10 and Recall@1, respectively. We hypothesize that this is due to the ``smearing" effect of averaging many different representations which leads to more robust document representations generally, but possibly at the cost of the high precision resulting from some referrals being an almost-perfect match for some queries at evaluation time. We conclude that for the retrieval task, concatenation for sparse models and mean for dense models results in the best overall performance, and use this configuration when reporting the main results in Table \ref{table-lcr}.

\begin{table*}
\centering
\begin{tabular}{lccc} \toprule
   & \multicolumn{1}{p{2cm}}{\centering \textit{Recall@1}}
   & \multicolumn{1}{p{2cm}}{\centering \textit{MRR@10}}
   & \multicolumn{1}{p{2cm}}{\centering \textit{Recall@10}} \\ \toprule
   BM25 & 0.115 & 0.157 & 0.265 \\
   \quad + referrals \textsubscript{concat} & \textbf{0.200} & \textbf{0.2677} & \textbf{0.505} \\
   \quad + referrals \textsubscript{sp} & 0.093 & 0.1406 & 0.255 \\
   \midrule
   SimCSE & 0.065 & 0.0869 & 0.160 \\ 
   \quad + referrals \textsubscript{concat} & 0.060 & 0.0989 & 0.190 \\ 
   \quad + referrals \textsubscript{mean} & 0.000 & 0.111 & \textbf{0.355}  \\ 
   \quad + referrals \textsubscript{sp} & \textbf{0.115} & \textbf{0.158} & 0.265  \\ 
   \bottomrule
\end{tabular}
\caption{\label{table-aggregation} Paper retrieval results, comparing different referral aggregation methods. We find that concatenation works best for the sparse model BM25, while mean works well for the dense model SimCSE and shortest-path achieves the best top-1 performance for SimCSE.}
\end{table*}

\subsection{Referrals allow for training-free modifications to the representation space}

One advantage of retriever models over large knowledge-base-like language models is the ability to easily add, remove, and otherwise update documents at inference time with no further fine-tuning. While knowledge editing and patching is an active area of research for large language models \cite{rome, editor}, all state of the art methods require costly optimization and remain far from matching the convenience and precision of updating a retriever-mediated information store, one reason search engines still dominate the space of internet-scale information organization.

We suggest that referrals naturally extend this property of retrievers, allowing not just documents but \textit{the conceptual relations between documents} and thus the \textit{effective representation space} to be updated without optimization. On top of adding newly available documents to a retrieval index, we can add their hyperlinks and citations to our collection of referrals, which not only improves retrieval performance on new documents but also \textit{continually improves the representations of older documents} with knowledge of new trends and structure.

\begin{table}
\centering
\begin{tabular}{lcc} \toprule
   & \multicolumn{1}{p{2cm}}{\centering ACL} \\ \toprule
   SimCSE & 0.325 \\ 
   \quad + referrals (up to 2018) & 0.615 \\
   \quad + referrals (up to 2019) & \textbf{0.665} \\
   \bottomrule
\end{tabular}
\caption{\label{table-more-years} Paper retrieval on 2020 papers with different referral cutoff years (Recall@10). We find that an updated referral pool improves referral-augmented retrieval.}
\end{table}

To demonstrate the impact of this in a realistic setting, in Table \ref{table-more-years} we show the improvement of SimCSE on paper retrieval (evaluating on queries constructed from papers published in 2020) when given additional referrals collected from the metadata of ACL papers released in 2019, compared to only referrals from papers up to 2018.\footnote{Specifically, we add the in-text citations of later layers to the pool of referrals, from which we randomly resample up to $\ell=30$ per document when building the retrieval index; the total number of citations is unchanged for most documents that already have 30 referrals available from the original dataset.} We see that augmenting from an updated pool of referrals improves performance by a significant margin.

Beyond adapting to newly available documents, referrals also open up the possibility of modifying document relationships for a variety of applications. \textbf{Human-in-the-loop corrections or additions} can be immediately taken into account by adding them as gold referrals, including adjusting a retrieval system to take trending keywords into account without changing the underlying document content. \textbf{Personalized referrals} such as mapping "favorite movie" to "Everything Everywhere All At Once" can also be recorded as a user-specific referral and can be updated at any time. Similarly, \textbf{temporary relations} for frequently changing labels such as the ``channel of the top trending video on YouTube" or ``Prime Minister of the UK" can be kept up to date using referrals. Clearly, we find that referrals unlock new abilities for retrieval systems beyond general improvements to performance.

\section{Related Work}

\paragraph{Sparse and dense retrieval} Following the success of BERT \cite{bert}, a variety of BERT-based dense encoder models have been proposed for information retrieval. \citet{dpr} propose DPR, fine-tuning on query-document pairs from MS MARCO \cite{msmarco}; \citet{simcse} propose SimCSE, fine-tuning using supervision from NLI datasets with entailment pairs as positives and contradiction pairs as hard negatives; and \citet{contriever} propose Contriever, fine-tuning using random crops and MoCo \cite{moco} to scale to a large number of negatives. However, \citet{beir} show that term-frequency sparse methods like BM25 remain a strong baseline in the zero-shot IR setting.

\paragraph{Query and document expansion} Query expansion techniques were originally proposed to decrease the lexical gap between queries and documents, using relevance feedback as well as external knowledge banks like WordNet \cite{wordnet}, whereas document expansion techniques such as Doc2Query and DocT5Query \cite{doc2query} were intended to add additional context and surface key terms.
Some work also explores sparse retrievers with learned document term weights \cite{splade} and late interaction models \cite{colbert}, which can be seen as performing implicit document expansion. However, most state-of-the-art dense retrievers \cite{simcse, dpr} do not perform any expansion, and in this work we have shown that they benefit significantly from referrals.

\paragraph{Hyperlinks for retrieval} Hyperlinks have been explored for use as a retrieval-first pre-training objective. \citet{harp} explore pre-training using the anchor text portion of a linking sentence as a pseudo-query for query-document pre-training, among other pre-training objectives, and \citet{php} improves upon this by defining different kinds of relevance classes based on where the hyperlink occurs and whether a pair of documents mutually link to each other, and performing multi-stage pre-training on (anchor text, linked document) pairs of increasing relevance. In contrast, we focus on using hyperlinks as training-free document augmentations to improve an arbitrary given encoder.

\paragraph{Citations for retrieval} Local citation recommendation is the task of retrieving a paper given a passage that cites it, and state-of-the-art approaches fine-tune using (citing paper’s title + abstract + citing passage, cited paper’s title + abstract) pairs \cite{lcr_nianlong}. Similarly, global citation recommendation is the task of retrieving relevant papers given a query paper, and state-of-the-art approaches include SPECTER \cite{specter}, in which fine-tuning is done on (citing paper’s title + abstract, cited paper’s title + abstract) pairs. Again, we focus on using citations as training-free referrals, and explore fine-tuning using pairs of single citing sentences that refer to the same paper. We notice that different citing sentences are often very similar, much more so than the titles and abstracts of pairs of citing and cited papers, leading to a cleaner supervision signal compared to passages and abstracts especially for the referral-aware setting.

\paragraph{Model updating and editing} An ongoing line of work \cite{rome, editor} studies fact editing for language models, which are resource-intensive to modify and trained on data that quickly becomes outdated. Retrieval systems trivially admit document edits and the addition of new documents without training, and we have found that hard negatives and referrals extend this property to support multiple document views. These benefits can reach end-to-end generation via retriever-augmented language models \cite{ic_ralm, realm}.

\section{Conclusion}

We propose a simple method to capture implicit hard positives using intra-document citations and hyperlinks as \textit{referrals} to provide alternate views of a given document, and show that referral augmentation yields strong model- and task-agnostic gains for zero-shot retrieval that outperforms previous text expansion techniques while also being less expensive. We also explore applications of hard positives as training-free modifications to the representation space, allowing new views of documents to be dynamically added to reflect updated world context, human-in-the-loop corrections, and personalized and temporary labels for documents.

One perspective on our referral augmentation results is evidence that an index that incorporates multiple views per document may be better suited for the retrieval of high-quality, atomic documents that may nevertheless each be relevant to a variety of different situations. It is also apparent that often these views may not be apparent from the document text itself — for example, a paper may be commonly referenced as the progenitor of a follow-up work, of which it obviously has no knowledge. Our work presents a preliminary look at a simple way to collect some of these nonobvious multiple views from the corpus itself, as well as the aggregation problem that subsequently arises; our work thus suggests that the more general problem of fully capturing these distinct facets of each document, and efficiently determining which facet is most relevant to a given query in a multi-view retrieval scenario, may be an important next step for robust retrieval.

\section*{Limitations}
The main limitation is that document-to-document links are not always available: referrals can be used with corpora such as academic papers and web-based articles, but not books or social media conversations.

We also note that the concatenation and shortest path aggregation methods lead to longer and more documents, respectively, in linear fashion in $\ell$, the number of referrals per augmented document. Thus, the augmentation trades off memory and speed for more relevant retrieved documents. This is tractable (and insignificant compared to the costs of generative expansion methods) with our choice of $\ell=30$ and fast max inner product search algorithms, but does impose a soft upper bound on the number of referrals it is feasible to take into account, especially for highly cited and linked documents. 

\section*{Ethics Statement}
The authors foresee no ethical concerns with the research presented in this paper.

\bibliography{anthology,custom}
\bibliographystyle{acl_natbib}

\end{document}